\documentclass[runningheads]{llncs}

\usepackage[utf8]{inputenc}
\usepackage{graphicx}
\usepackage{siunitx}
\usepackage{hyperref}
\usepackage{xspace}
\usepackage{amsmath, amsfonts, amssymb}
\usepackage{bm}
\usepackage{subcaption}

\hypersetup{
    colorlinks=true,
    linkcolor=blue,
    urlcolor=blue
}

\DeclareMathOperator*{\argmin}{arg\,min}
\newcommand*{\eg}{e.\,g.\@\xspace}
\newcommand*{\ie}{i.\,e.\@\xspace}

\newcommand*{\etc}{\@ifnextchar{.}{etc}{etc.\@\xspace}}

\begin{document}

\title{Vectorial Genetic Programming -- Optimizing Segments for Feature Extraction}
\titlerunning{Vectorial GP -- Optimizing Segments for Feature Extraction}

\author{
	Philipp Fleck\inst{1,2}\thanks{corresponding author\\ \email{philipp.fleck@fh-hagenberg.at}}\orcidID{0000-0001-8290-6356}
	\and
	Stephan Winkler\inst{1,2}\orcidID{0000-0002-5196-4294}
	\and
	Michael Kommenda\inst{1,3}\orcidID{0000-003-2049-723X}
	\and
	Michael Affenzeller\inst{1,2}\orcidID{0000-0001-5692-5940}
}

\authorrunning{P. Fleck et al.}

\institute{
	Heuristic and Evolutionary Algorithms Laboratory (HEAL),\\
	{\scriptsize University of Applied Sciences Upper Austria, Softwarepark 11, 4232 Hagenberg, Austria}
	\and
	Institute for Symbolic Artificial Intelligence,\\
	{\scriptsize Johannes Kepler University, Altenberger Straße 69, 4040 Linz, Austria}
	\and
	Josef Ressel Center for Symbolic Regression,\\
	{\scriptsize University of Applied Sciences Upper Austria, Softwarepark 11, 4232 Hagenberg, Austria}
}

\maketitle

\begin{abstract}
Vectorial Genetic Programming (Vec-GP) extends GP by allowing vectors as input features along regular, scalar features, using them by applying arithmetic operations component-wise or aggregating vectors into scalars by some aggregation function.
Vec-GP also allows aggregating vectors only over a limited segment of the vector instead of the whole vector, which offers great potential but also introduces new parameters that GP has to optimize.
This paper formalizes an optimization problem to analyze different strategies for optimizing a window for aggregation functions.
Different strategies are presented, included random and guided sampling, where the latter leverages information from an approximated gradient.
Those strategies can be applied as a simple optimization algorithm, which itself ca be applied inside a specialized mutation operator within GP.
The presented results indicate, that the different random sampling strategies do not impact the overall algorithm performance significantly, and that the guided strategies suffer from becoming stuck in local optima.
However, results also indicate, that there is still potential in discovering more efficient algorithms that could outperform the presented strategies.

\end{abstract}

\section{Introduction}

\begin{figure}
	\centering
	\includegraphics[width=0.8\linewidth]{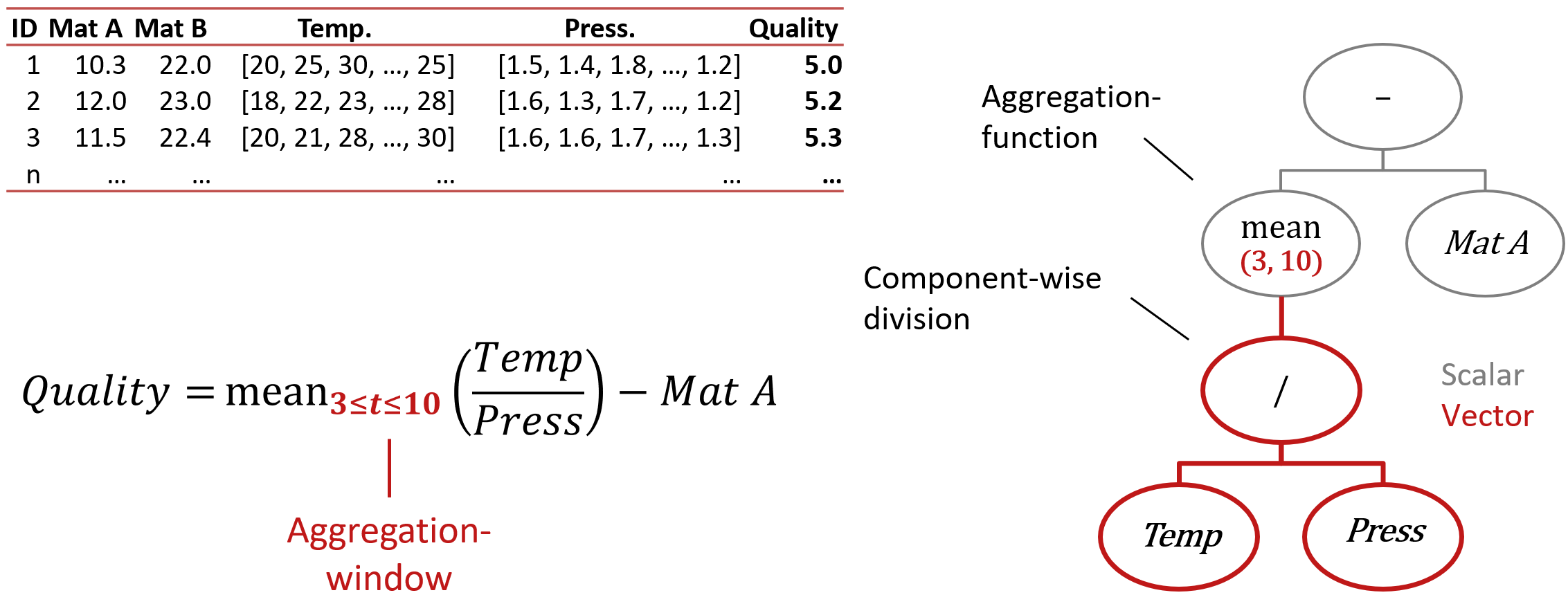}
	\caption{An example dataset containing both scalar and vector input features to predict a scalar target value (the quality). It also shows a GP model that uses a windowed-aggregation function to aggregate a vector into a scalar.}
	\label{fig:vec_GP_aggr_windows}
\end{figure}

Vectorial Genetic Programming (Vec-GP) for Symbolic Regression (SR) is a extension of GP, where the space of input features is extended to vectors alongside regular scalars\cite{Azzali2019}.
This allows vectorial GP to directly handle higher dimensional data, such as time series, without the need of prior feature engineering to extract scalar features.
Instead, vectorial GP models themselves extracts scalar values from the vectors by applying aggregation functions, such as the arithmetic mean.
Compared to traditional feature engineering, where scalar features are usually only extracted from the raw vector features, vectorial GP can also extract features from interacting vectors by performing arithmetic operations on vectors before aggregation takes place.
Fig.~\ref{fig:vec_GP_aggr_windows} shows an example, where the vector variable for temperature and pressure are divided prior to calculating the arithmetic mean to obtain a scalar.
The literature already indicates, that vectorial GP outperforms regular GP with feature engineering in various benchmarks\cite{Fleck2021}.

In case of a vector representing a time series, it might be beneficial to aggregate only over a certain time frame.
In this sense, vectorial GP can be further extended by adding the ability to aggregate not only over a whole vector, but over a sub-segment of the vector, where GP is also in charge of optimizing the segment bounds.
For instance, as also shown in Fig.~\ref{fig:vec_GP_aggr_windows}, aggregation might only be done on a specific time frame, for example, between $3 \le t \le 10$.

\section{Problem Description}

When using an aggregation function with an aggregation window, GP is now required to optimize two additional parameters for each windowed aggregation function within a GP model.
While GP could handle this by allowing the indices to change via mutation, random mutation is not particularly well suited for this task, especially for longer vectors.
Instead of relying on mutation as GP's only way of optimizing aggregation windows, this paper is a first step towards identifying other strategies that are more tailored towards optimizing aggregation windows, similar to applying efficient  gradient-based algorithms for optimizing continuous, numerical parameters of a fixed GP model\cite{kommenda2020parameter}.


To focus on the problem of optimizing the aggregation window, we use a simplified optimization problem that omits any GP related aspect, only keeping the core problem of optimizing an aggregation window over some fixed data.
This Segment Optimization Problem (SOP) is an integer problem, optimizing the parameters $\hat{a}, \hat{b} \in \{ i \in \mathbb{N} \mid i < N \}$ representing the start and end index with, minimizing
\begin{equation}
	\argmin_{\hat{a}, \hat{b}}
		\sum_{i=0}^{M}\Big(
			\underbrace{
				\mathrm{f}_\mathrm{agg}\big( \boldsymbol{v_i}[\hat{a}:\hat{b}] \big)
			}_{\text{optimized bounds}}
			-
			\underbrace{
				\mathrm{f}_\mathrm{agg}\big( \boldsymbol{v_i}[a:b] \big)
			}_{\text{known bounds}}
		\Big)^2,
\end{equation}
where $\mathrm{f}_\mathrm{agg}$ is a fixed aggregation function, $\boldsymbol{v_i} \in \mathbb{R}^{M \times N}$ are the $M$ samples of vectors with length $N$, $a$ and $b$ representing the known aggregation indices and $\boldsymbol{v_i}[a:b]$ being the sub-vector of a selected vector $\boldsymbol{v_i}$.
In essence, the goal of the SOP is to find the minimum deviation when aggregating using the free indices $\hat{a}$ and $\hat{b}$ compared to aggregating using the known indices $a$ and $b$.

\section{Method}
\label{sec:Method}

Instead of relying on GPs mutation to identify good aggregation windows, our goal is to define a specialized mutation operator for GP that is tailored towards optimizing the start and end indices of an aggregation window efficiently.
Start and end indices can either be optimized simultaneously, \ie as a multi-dimensional optimization problem, or only a single dimension at a time, \ie as a single-dimensional optimization problem.
We will later analyze which form yields better results.

\begin{figure}[b]
	\centering
	\includegraphics[width=0.7\linewidth]{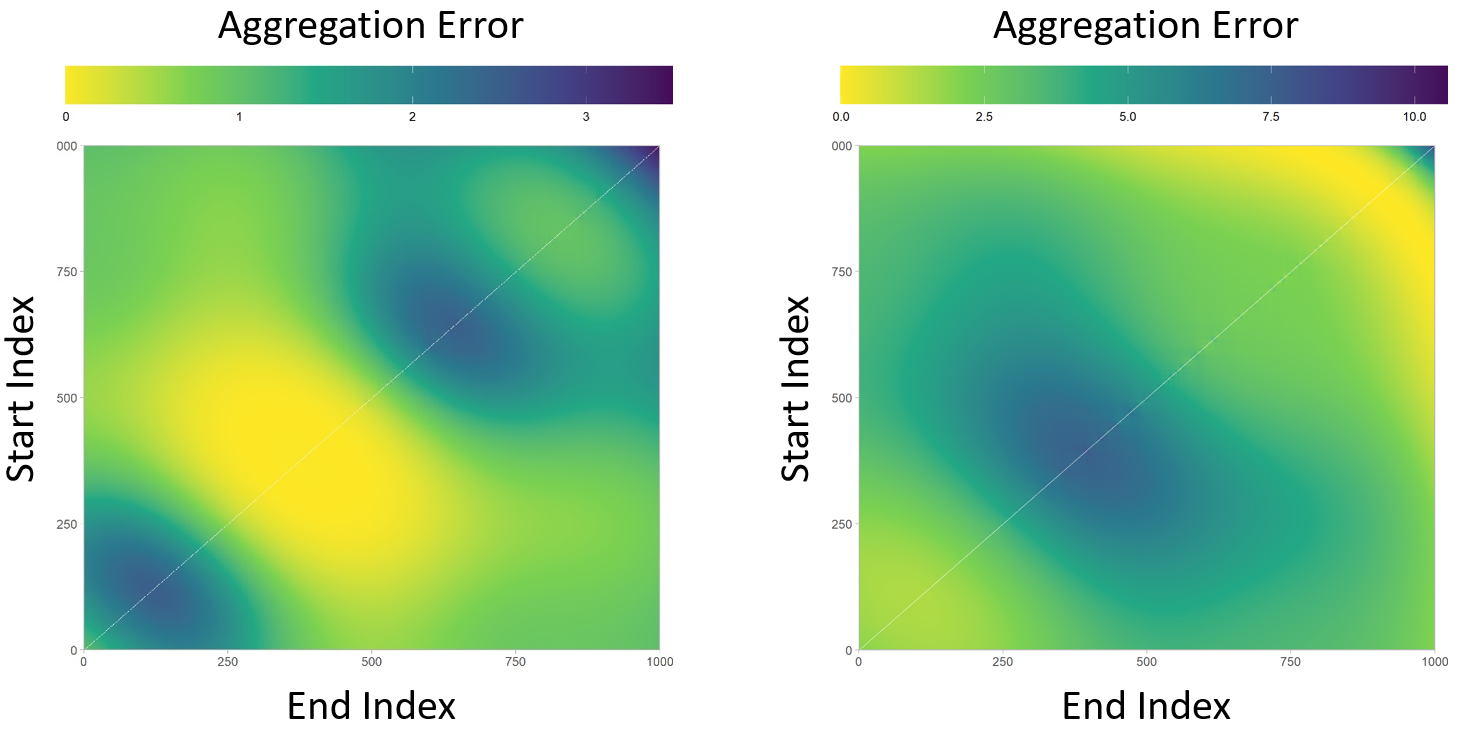}
	\caption{The fitness landscape of two benchmark instances (x3 and x6) with lower (brighter) values indicating better regions in the search space.}
	\label{fig:fitness_landscape}
\end{figure}

Looking at the fitness landscape of some of the benchmark instances in Fig.~\ref{fig:fitness_landscape}, we can clearly observe, that there is gradient information that could be leveraged for efficient optimization.
Therefore, we first approximate the gradient by using a five-point stencil\cite{sauer2012numerical} on the neighboring indices ($h=1$) and then use it to guide the search.
The proposed optimization method in this paper is an iterative algorithm with two steps:
First, the gradient is calculated and used to guide search towards promising regions.
Second, potential solutions are sampled from that promising region, evaluated and the best one is selected for the next iteration.
This procedure is continued until a predefined number of total solution evaluations is exceeded.

\begin{figure}
	\centering
	\includegraphics[width=0.7\linewidth]{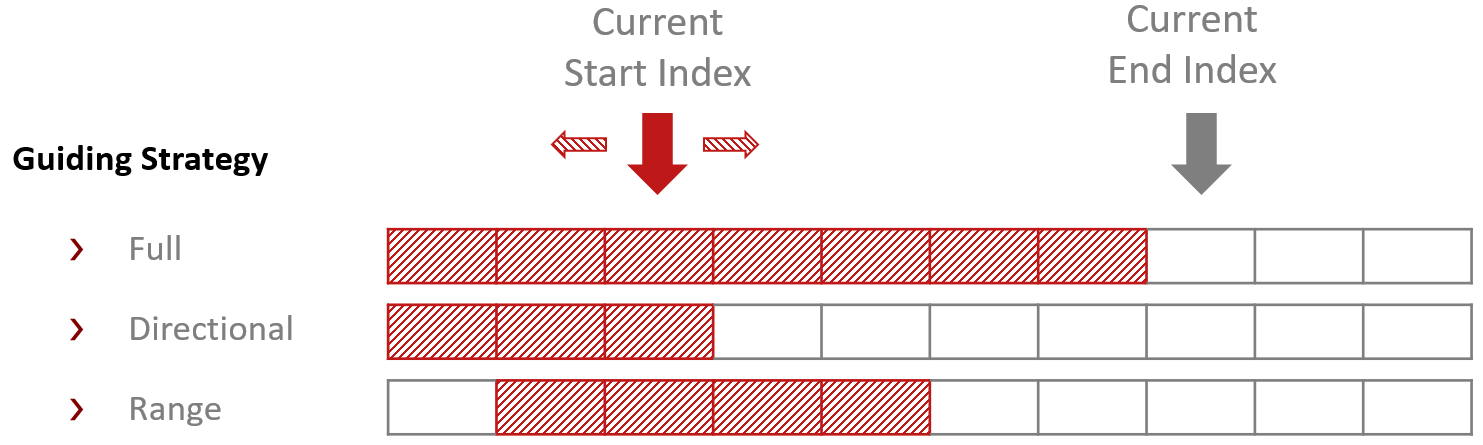}
	\caption{Three different guiding strategies for defining the search space, given a free start index and a fixed end index.}
	\label{fig:GuidingStrategy}
\end{figure}

The firsts guiding step in the optimization procedure is responsible for selecting a promising region, from which samples can be drawn in the second step afterwards.
We define three different guiding strategies, shown in Fig.~\ref{fig:GuidingStrategy}:
\emph{Full} represents no strategy at all, where the whole solution space is utilized. This strategy serves will serve as a baseline.
\emph{Direction} uses the approximate gradient only to guide whether an index should be increased or decreased and does not take the magnitude of the gradient into account.
\emph{Range} uses the approximate gradient information to define a promising region by a defined range, where the magnitude of the gradient also influences the distance from the current index.

\begin{figure}[tbh]
	\centering
	\includegraphics[width=0.70\linewidth]{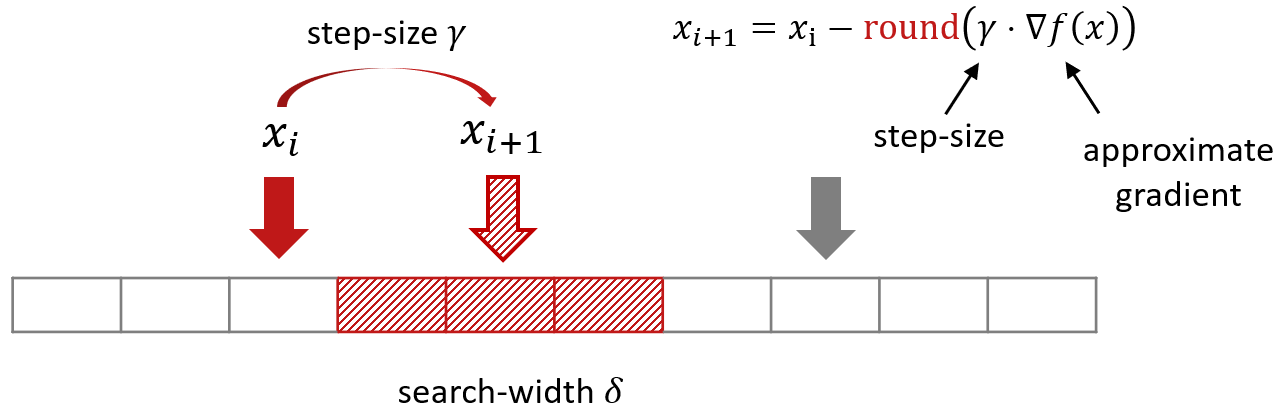}
	\caption{Adoption of a gradient descent step to integer indices, with an additional search-range parameter.}
	\label{fig:GuidedRange}
\end{figure}

While the random direction simply looks at the direction of gradient, the guided range mimics a traditional gradient descent algorithm with two modifications, shown in Fig.~\ref{fig:GuidedRange}.
First, we round the gradient step towards the nearest integer, to ensure that the next index will also be valid index.
Additionally, we introduce a new parameter, the search-range, that defines how many samples around the next index are considered promising and are thus eligible for being sampled afterwards.

\begin{figure}[tbh]
	\centering
	\includegraphics[width=0.7\linewidth]{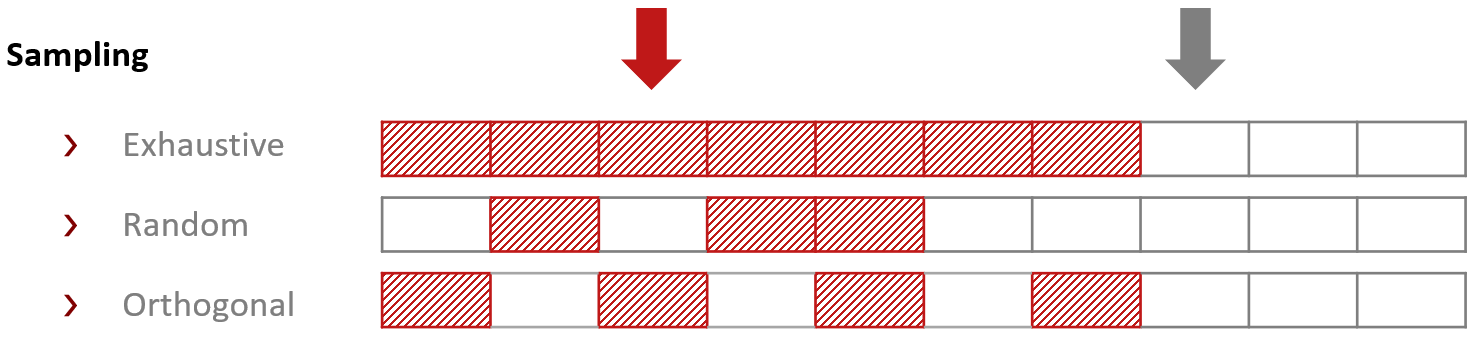}
	\caption{Different sampling strategies for a given search space.}
	\label{fig:SamplingStrategy}
\end{figure}

After defining the search space, the second step is responsible of drawing the samples from this search space.
For this paper, we have considered three sampling mechanisms, shown in Fig.~\ref{fig:SamplingStrategy}:
\emph{Exhaustive} draws all samples from the solution space. This is typically not feasible for a multi-dimensional search space, due to a potentially high number of possible solutions.
However, it could be a viable option for a single dimensional space.
\emph{Random} draws samples completely at random (without repetition), where the number of samples are defined by an additional sample-size parameter.
\emph{Orthogonal} selects the samples equally spaced along each dimension, by first selecting a predefined number of equally spaced points on a continuous line of the search space and then rounding towards the nearest indices and removing any duplicates.

\section{Experiment Setup}
\label{sec:Experiment_Setup}

To identify good parameters of the search strategies defined in the previous section, we performed a grid search of the parameters over multiple benchmark instances.
However, certain combinations were skipped due to infeasible runtimes, \eg exhaustive search on multi-dimensional search spaces.
Each parameter combination was repeated 20 times, and each execution of the optimization algorithm was allowed a total of 100,000 sample evaluations, meaning the total number of iterations varied, depending on the sample-size, for instance.

\begin{figure}[tbh]
	\centering
	\includegraphics[width=1.0\linewidth]{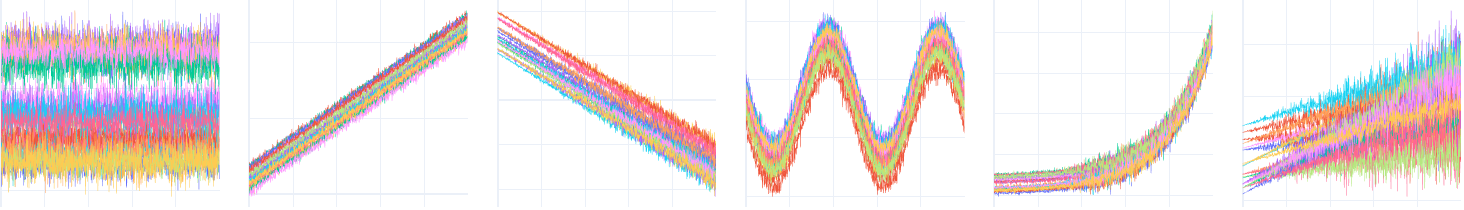}
	\caption{Some exemplary benchmark instances, where each line represent one row in the dataset.}
	\label{fig:BenchmarkInstances}
\end{figure}

The generated benchmark instances have different characteristics to represent different challenges and difficulties.
For instance, shown in Fig.~\ref{fig:BenchmarkInstances}, the second instance shows an overall increasing behavior where the noise stays constant.
The most right benchmark, however, does not only have more varying slopes, but also the noise is increasing over time.
The benchmark instances were created in multiple steps by creating randomized control values first, that are then used as arguments for other distributions that are finally used to create the vectors.
For instance, the most right benchmark is based on the control values $m = \mathcal{U}(2, 5) + \mathcal{U}(0.2, 4) / 80 \cdot t$ and $s = \mathcal{U}(0, 0.001) + \mathcal{U}(0.002,0.01) \cdot t$ where $\mathcal{U}$ denotes a uniform random variable and $t$ denotes the time or index.
The final vector is then created using a normal distribution $\mathcal{N}(m, s^2)$ with the prior defined control values.
Thus, for this benchmark instance, the mean value and also the noise increases with higher $t$.
Since the parameters of the distributions change with $t$, the distributions from which a sample is drawn are created for each $t$ independently.
The full list and definition of the benchmark instances can found in the additional materials online at \url{https://dev.heuristiclab.com/wiki/AdditionalMaterial}.
For the presented results, we used vectors of length 1,000.

\section{Result}

To compare different optimization strategies, we analyze the run-length distributions, as defined in the black-box optimization benchmarking (BBOB)\cite{hansen2010comparing}.
This allows an evaluation of not only an algorithms ability of finding a good solution, but also shows how fast an algorithm converges.
This is crucial, since a good optimization strategy for the segment optimization problem would be used inside a mutation operator within GP, thus, long runtimes would multiply in the context of GP.

\begin{figure}
	\centering
	\includegraphics[width=0.7\linewidth]{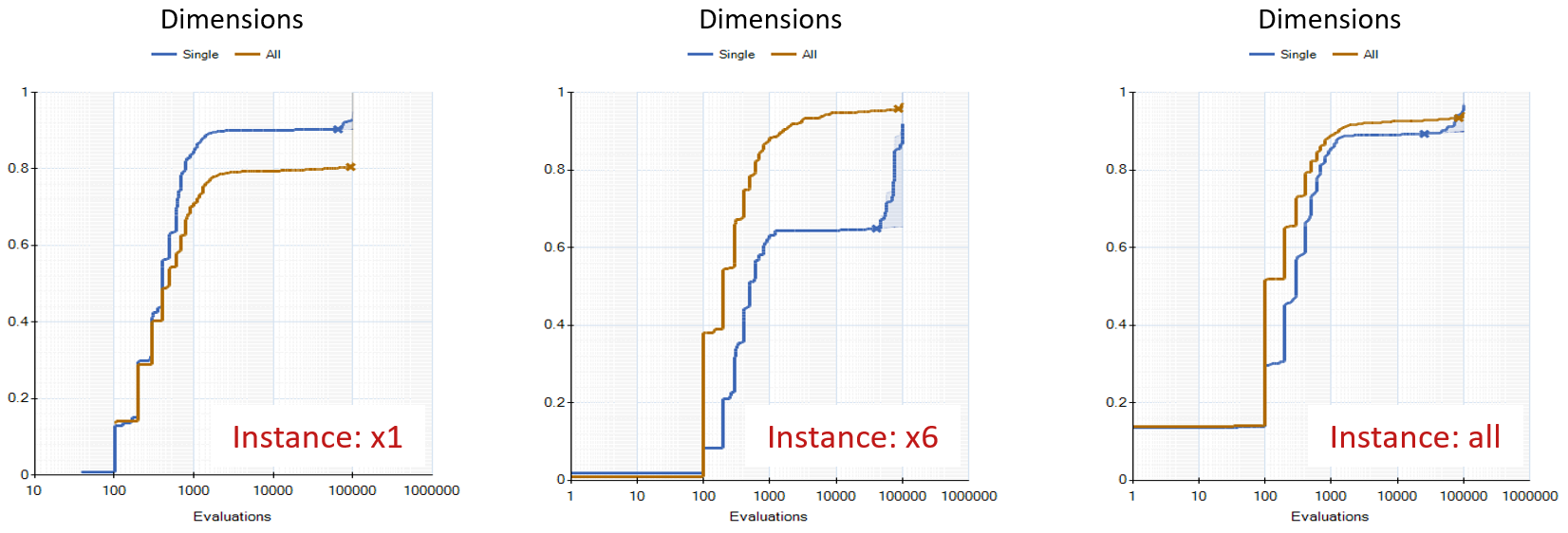}
	\caption{Influence of single- versus multi-dimension on algorithm performance.}
	\label{fig:dimension_comparison}
\end{figure}

In addition to the presented results, we also analyzed additional parameters, such as the effect of random versus orthogonal sampling, however they showed no significant differences, thus are not discussed further in this paper.
We will first look at whether approaching the optimization problem as single- or multi-dimensional performs better, shown in Fig.~\ref{fig:dimension_comparison}.
Overall, the results suggest, that optimizing all dimensions simultaneously performs slightly better, however, for some problem instances, such as x1, using a single dimension yielded better results.
This behavior was suspected, since it shows, that some of the benchmark are easy enough to be separable and optimized dimension by dimension.

\begin{figure}
	\centering
	\includegraphics[width=0.7\linewidth]{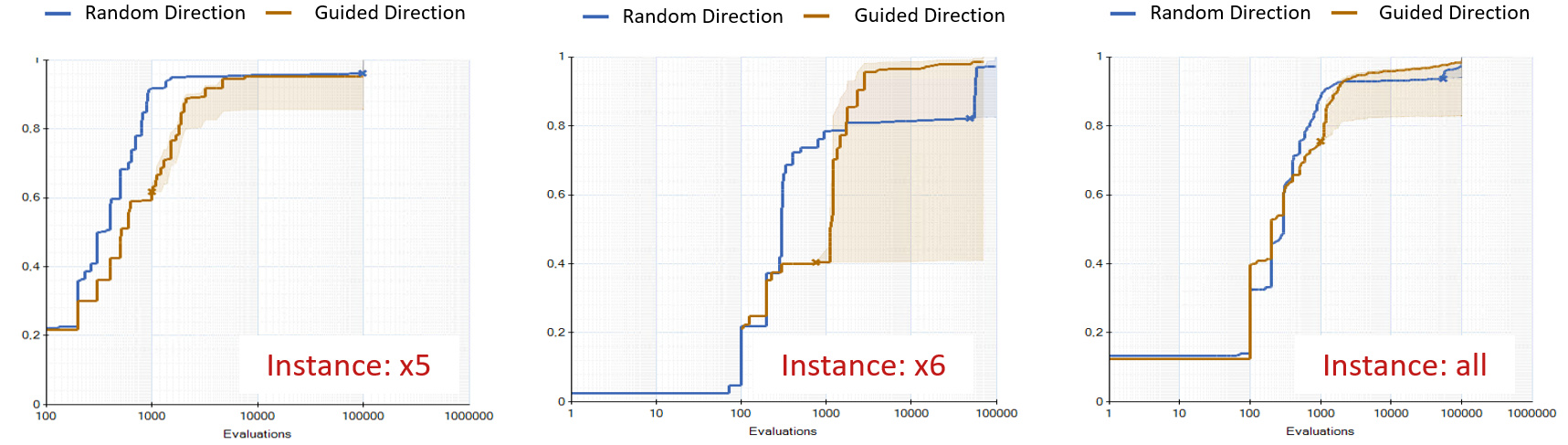}
	\caption{Performance of random versus guided direction.}
	\label{fig:guided_vs_random_direction}
\end{figure}

Next, we analyze directional search space reduction, where we compare limiting the search space towards a direction randomly versus using the gradient to guide the direction, shown in Fig.~\ref{fig:guided_vs_random_direction}.
The result suggests, that the benefits depend on the problem instance.
While for problem instance x5 using a guided direction seems to slow down convergence, overall, the benefits are not conclusive.
We suspect, that by using the gradient always to move towards the best option, it forces the algorithm towards local optima without any way of escaping.

\begin{figure}
	\centering
	\begin{subfigure}{0.49\textwidth}
		\centering
		\includegraphics[width=\textwidth]{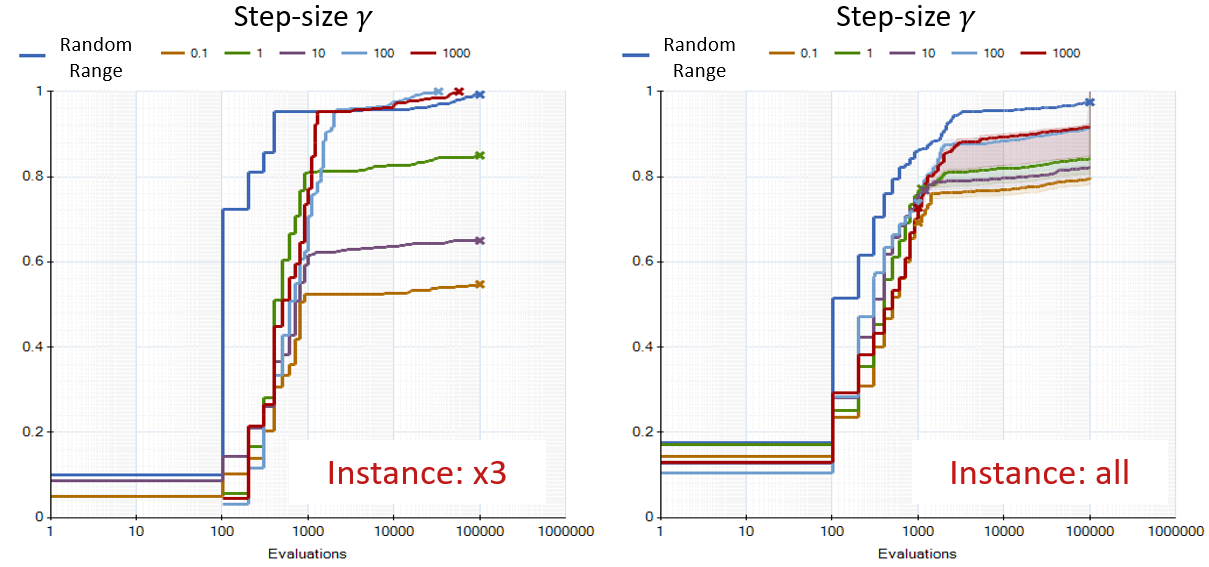}
		\caption{Influence of the step-size.}
		\label{fig:step_size_comparison}
	\end{subfigure}
	\begin{subfigure}{0.49\textwidth}
		\centering
		\includegraphics[width=\textwidth]{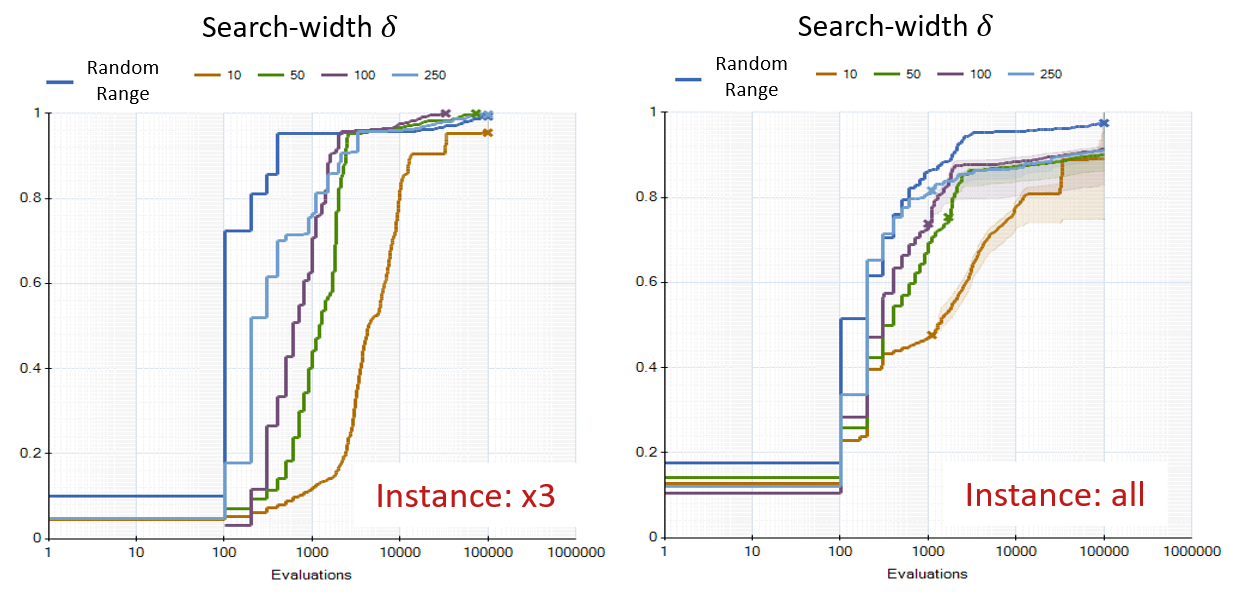}
		\caption{Influence of the search-width.}
		\label{fig:search_width_comparison}
	\end{subfigure}
	\caption{Performance of random versus guided search range reduction.}
	\label{fig:guided_vs_random_range}
\end{figure}

Next, we analyze limiting the search space within a specific range.
Because the range strategy has two additional parameter, the step-size and search-width, we will compare the random version to different settings of the parameters, both shown in Fig.~\ref{fig:guided_vs_random_range}.
Since the step-size controls the jumping distance from the current index, having low values are expected to yield bad results, since jumping too little makes the search inefficient or halt completely due to the rounding in the gradient step.
The effect of the step-size on instance x3 suggest that this assumption holds, since lower step-sizes do not perform well, as shown in Fig.~\ref{fig:step_size_comparison}
On the other hand, higher step-sizes tend to perform better for instance x3 and might even outperform the random range strategy in finding the best solutions quicker, however, this is not the case for all instances.
Along the step-size, the search-width defines how far the search space extends around the next index.
The results in Fig.~\ref{fig:search_width_comparison} suggests, that having a lower search-width generally causes the algorithm to converge slower.
This is especially true for combinations with a low step-size, since the search-range is the only mechanism of changing an index when the gradient step is rounded to zero.
Similar to the guided direction, where we suspect that random direction outperforms the guided direction because the algorithm becomes stuck at a local optimum, the same could also be true for guided range to a certain degree.
Having a large search-width could help breaking out of the local optimum, but still, the result suggest that a random range would be the better option in most cases.

\section{Conclusion}

The results presented in the previous section suggests, that simple random sampling is generally better than using a guided strategy such as guided direction or guided range.
We believe that this is mainly caused by the guided strategies tend to work similar to greedy algorithms, and can quickly become stuck in local optima.
However, when applying the presented methods in the context of GP, preliminary results suggests, that the guided strategies tend to outperform their random counterparts.
We suspect that GP can easily break free of local optima by regular crossover and mutation operations, thus becoming stuck in a local optima is not as critical.

Despite the current result suggesting that the gradient-based strategies do not work well, we still believe that the information of the approximate gradient is extremely useful, because most fitness landscapes of the benchmark instances clearly indicate that there is a gradient that could be exploited for fast optimization.
Therefore, we will continue working on leveraging the gradient for a fast optimization algorithm for the segment optimization problem.

\subsubsection*{Acknowledgments}
This work was carried out within the Dissertationsprogramm der Fachhochschule OÖ \#875441 \emph{Vektor-basierte Genetische Programmierung für Symbolische Regression und Klassifikation mit Zeitreihen (SymRegZeit)}, funded by the Austrian Research Promotion Agency FFG.
The authors also gratefully acknowledge support by the Christian Doppler Research Association and the Federal Ministry of Digital and Economic Affairs within the \emph{Josef Ressel Centre for Symbolic Regression}.

\bibliographystyle{splncs04}
\bibliography{bibliography}

\end{document}